\documentclass[11pt]{article}
\usepackage{geometry}                
\usepackage{graphicx}
\usepackage{amssymb}
\usepackage{epstopdf}
\title{Comments on Sejnowski's ``The unreasonable effectiveness of deep learning in artificial intelligence" (2020).}
\author{Leslie S. Smith \\Computing Science and Mathematics\\ University of Stirling\\
Stirling FK9 4LA\\ Scotland, UK\\ email l.s.smith@cs.stir.ac.uk}

\begin{document}
\maketitle
\begin{abstract}
Terry Sejnowski's 2020 paper [arXiv:2002.04806] is entitled ``The unreasonable effectiveness of deep learning in artificial intelligence". However, the paper itself doesn't attempt to answer the implied question of why Deep Convolutional Neural Networks (DCNNs) can approximate so many of the mappings that they have been trained to model. While there are detailed mathematical analyses,  this short paper attempts to look at the issue differently, considering the way that these networks are used, the subset of these functions that can be achieved by training (starting from some location in the original function space), as well as the functions to which such networks will actually be applied.
\end{abstract}

Terry Sejnowski's paper is entitled {\em The unreasonable effectiveness of deep learning in artificial intelligence}, \cite{Sejnowski:2020el}. and he compares and contrasts deep learning with Wigner \cite{Wigner:1960ue} who marvelled at the limited numbers of parameters in equations. This is in  contrast with modern deep learning with its abundance of parameters and extremely high dimensional spaces.  While he discusses how we should look to the brain for further optimization, in the paper he does not attempt to answer the question of {\em why}  deep learning is so unreasonably effective.  So why can Deep Convolutional Neural Networks (DCNNs)  approximate so many of the mappings that they have been trained to model?

There are detailed mathematical analyses, for example \cite{Hornik:1991kg} \cite{Lin:2017cm}, \cite{Grohs:2019wr} (and the many references within  these papers), but this short paper attempts to look at the issue differently, considering the way that these networks are used,  the subset of these functions that can be achieved by training (starting from some location in the original function space), as well as the functions that in reality will be modelled.

Deep neural networks have a great many parameters. Firstly, there is the architecture of the network, constrained only by the number of inputs and outputs, and then there are the actual (trainable) parameters of the network itself. For a deep feedforward network, the number of these is set by the number of layers, and  by number of pseudo-neurons (hereafter neurons) in each layer. For DCNNs, it also depends on how the convolutions are implemented. The activation function and output function of each neuron is a parameter, though these are not usually trained. The original topology is also constrained both on the nature of the mapping being approximated, and by how the inputs and outputs for the network have been coded, but these, and the actual internal topology (number of layers, number of neurons per layer, how and where convolution is performed) are not normally optimized during network training: only the weights and biasses are changed. 

We write
$\cal{F}$ for the set of functions that the network can perform, and $f^1_{\rm init} \in \cal{F}$ for the function that is implemented by the network prior to training. Training will lead to a sequence of functions $f^i_{\rm init} (i \geq 1)$ eventually converging (assuming that learning is set up so that it does converge) to some $f^{\rm final}_{\rm init} \in \cal{F}$. Clearly, the sequence of functions will depend on $f^1_{\rm init}$, on the actual topology, and on the dataset and precise learning and adaptation rules used to train the network.


Networks are generally trained using a dataset $D = (D_{\rm in}, D_{\rm out})$, where $D_{\rm in} = \{d^j_{\rm in}: j = 1 \ldots t\}$, and $D_{\rm out} = \{d^j_{\rm out}: j = 1 \ldots t\}$. $D$ is thought of as sampling some underlying classification or function $f_D$. $t$ is the total number of training examples. $d^j_{\rm in}$ is a  vector whose length is the size of the input layer, $N_{\rm input}$, and  $d^j_{\rm out}$ a vector whose length is the size of the output layer, $N_{\rm output}$. The aim is to be able to map other unseen inputs $d^{\rm new}_{\rm in}$ to appropriate $d^{\rm new}_{\rm out}$. For classification, there will be a $j$ such that $d^{\rm new}_{\rm out} = d^{\rm i}_{\rm out}$, but for function approximation, $d^{\rm new}_{\rm out}$ may be novel.

For a particular training dataset $D$, we will normally {\em choose} an initial topology $\tau = \tau(D)$ that is likely to be able to solve the problem: that is, for which there is likely to be an 
\begin{equation}
f^{\rm final}_{\rm init} = f^{\rm final}_{\rm init}(\tau(D)) 
\label{eqn:1}
\end{equation}
which approximates the characteristics of $D$ within some range of error $\epsilon$\footnote{We are attempting to be general here: the nature of $D$, and the nature of $\epsilon$ will depend on the problem at hand.}. We note that if experiment shows that this is not the case, we can choose a different topology $\tau'$ until we find one for which this is true. 

We  note that $D$ is not arbitrary,  but has some real-world problem at its root. Thus (for example) we would not be trying to learn a completely random classification of $N_{\rm input}$ binary inputs (which would essentially require $2^N$ parameters to learn precisely, or rather fewer than that if we choose a nonzero $\epsilon$). But can we say anything else about $D$, or about the classification or function that $D$ is a sample from (which is more important, since it is generalisation that we are really interested in)? In practice, $D$ is a sample from some underlying function $f_D$, mapping a set of inputs to a set of outputs, rather than a collection of arbitrary input/output pairs, and it is  $f_D$ that we are trying to approximate.   In \cite{Grohs:2019wr} there is some discussion of the variety of smooth functions for which deep neural networks are good approximators.

The function space $\cal{F}$  (which $f_D$ is a member of) can be characterised as
\begin{equation}
 \cal{F} = {\mathbb R}^{N_{\rm input}} \times{\mathbb R}^{N_{\rm output}} 
 \end{equation}
 where $N_{\rm input}$ is the the number of input units and $N_{\rm output}$ is the number of output units. However, because computers are finite devices, the space is much smaller than this, more like
 \begin{equation}
F' =  (2^{64})^{N_{\rm input}} \times  (2^{64})^{N_{\rm output}}
 \end{equation}
 assuming that both inputs and outputs are coded in 64 bits (as is currently likely to be the case for 64 bit floating point coded inputs).

If we are successful in approximating $f_D$, this means that there is (and we have found) a $\tau$ such that  $f^{\rm final}_{\rm init}(\tau)$ is close enough to $f_D \in F'$. 
But what sorts of mappings can the network produce? This clearly depends on  the number of layers in the network, the way in which convolving layers are used, and (assuming the usual form of artificial neuron is used), the function used to compute the output from the activation of a neuron, and is the question  asked in \cite{Grohs:2019wr} .

Earlier we noted that $f_D$ was not arbitrary: but if we are to understand how $f^{\rm final}_{\rm init}(\tau)$ can be close enough to $f_D \in F'$ we need to refine what {\em not arbitrary} might mean. We identify two issues here: 
\begin{description}
\item[Issue 1] the nature of likely $f_D$'s, defined by the nature of the problem from  which $D$ was sampled
\item[Issue 2] the actual domain of the function that $D$ was drawn from
\end{description}

Consider an image classification problem: for such a problem, we start with images (that is, $d^i_{\rm in}$ is a coded image, probably an $X$ by $Y$ pixel image, with each pixel coded in 8 (monochrome) or 24 (colour) bits). These images result from incident light (from many sources, usually) reflecting from (and/or being transmitted by) points in the world, passing through the point spread function of each pixel detector at the camera, resulting in the $X$ by $Y$ vector that we are trying to interpret. 

\begin{figure}[htbp]
\begin{center}
\includegraphics[width=0.8\linewidth]{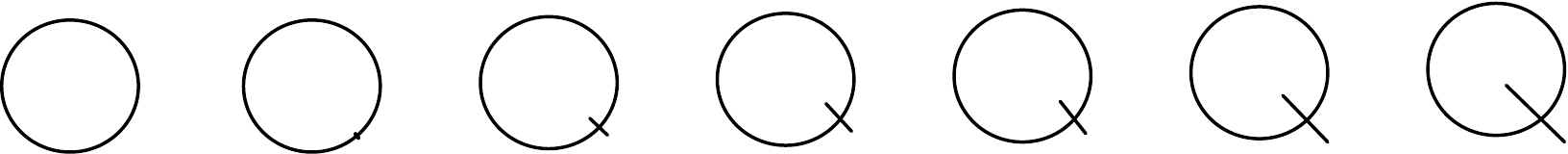}
\caption{When should an image classifier say that the letter O turns into a letter Q?}
\label{fig:OtoQ}
\end{center}
\end{figure}

Considering Issue 1 above, there are issues of size and translation invariance in $f_D$. In addition, we would expect the same classification for an image and a slightly blurred version of that image, and we would expect the classification to remain the same under a range of illuminations. Thus the $f_D$ from which $D$ is drawn is quite tightly constrained. This effect is not limited to image classification: something similar is true for sound classification, in that small alterations in the intensity or pitch of the sound, or in the reverberation of the sound prior to transduction would again not be expected to alter the classification of the sound. Where this becomes more difficult is in the case of categorical perception, as in figure \ref{fig:OtoQ}: each change is very small, but at some point one needs to change the classification.

\begin{figure}[htbp]
\begin{center}
\includegraphics[width=0.3\linewidth]{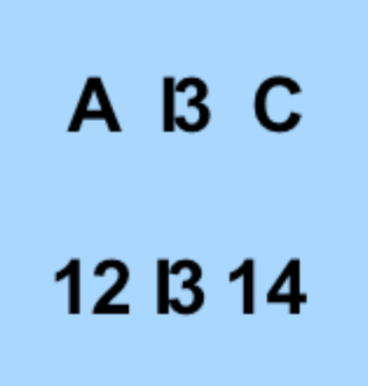}
\caption{When should an image classifier say that the middle symbol is B and when 13?}
\label{fig:Bto13}
\end{center}
\end{figure}

A related issue arises with context: see figure \ref{fig:Bto13} (from \cite{Kay:2018ww}). While the first issue may be soluble using a neural network, the second one requires the use of the context of the symbols on either side, as the central images are identical.

For Issue 2 above, one might at first think that any possible vector that is $X$ by $Y$  with 8 (or 24) bit elements might be a possible $d^i_{\rm in}$. However, this is generally not the case: in general, pixel values do not change suddenly or randomly between adjacent locations. If one considered $\cal{F}$ rather than $F'$, and considered smooth functions, one might be able to restrict it by bounding local partial derivatives. The overall effect is that although  $f_D$ is sampled from a function on the input space (whether $\mathbb{R}^{N_{\rm input}}$ or $ (2^{64})^{N_{\rm input}} $), many (indeed most) elements of this input space will never occur. Thus, testing $f^{\rm final}_{\rm init}(\tau)$ on randomly selected elements of the input space may be inappropriate, and lead to unexpected results  as in \cite{Szegedy:2013vwa}, as well as more or less random patterns being misclassified (when  they should be marked as unclassifiable).

What do these issues say to {\em The unreasonable effectiveness of deep learning in artificial intelligence}?

Firstly, we note that the {\em unreasonable effectiveness} is posited on the selection of an appropriate network $\tau(D)$ by the developer: while the mathematics might suggest that a single very wide hidden layer  network might suffice \cite{Hornik:1991kg}  the actual networks used do not look like this, primarily because appropriate generalisation is more important than being able to create a precise function.  The arguments in \cite{Lin:2017cm} and \cite{Grohs:2019wr} are unaffected, but it is clear that the space of functions (particularly classifiers) that require to be approximated is smaller than might have been imagined. There is a sense in which the functions are relatively smooth, and constrained further by the application domain. Further, the neural network will always classify or approximate any input from the input domain, whether that input is possible (in terms of the actual problem input domain) or not. 

Sejnowski discusses some unanswered questions, specifically why it is possible to generalize from so few examples while using so many parameters.  Based on what we have explained, because the classifier (or function approximator) is likely to include dimensionality reducing sections within the network (such as relatively narrow layers, or convolution layers), these inputs may be well away from the manifold that $f_D$ was actually trained on (and therefore gives appropriate results for), and may give inappropriate results. However, actual data from the same source as the training data will not suffer from this problem.   Thus we conclude that the unreasonable effectiveness while certainly true, is perhaps just a little less surprising.

Do the issues above affect the reachability of an appropriate $f^{\rm final}_{\rm init}(\tau)$? Issue 2 above suggests that testing of the function should be on problem-appropriate possible inputs, rather than from inputs drawn randomly from the domain. Thus  $\epsilon = \epsilon(f_D, \tau, f_{\rm init}^1)$ should not be evaluated on randomly selected elements of the input domain. This fits with many test problems, where a dataset is supplied, and is used for training and testing. It is however not clear that this affects reachability directly: however, the user selection of $\tau$ (and $ f_{\rm init}^1)$ is clearly important here.

\section*{Acknowledgement}
Thanks to Andrew Abel for useful comments on an earlier draft.

\bibliographystyle{halpha}
\bibliography{Sejnowski2020}

\end{document}